\pdfoutput=1

\documentclass[11pt]{article}

\usepackage{acl}

\usepackage{times}
\usepackage{latexsym}

\usepackage[T1]{fontenc}

\usepackage[utf8]{inputenc}

\usepackage{microtype}

\usepackage{svg}
\usepackage{graphicx}
\usepackage{multirow}
\usepackage{amsmath}
\usepackage[linesnumbered,ruled,vlined]{algorithm2e}
\usepackage{adjustbox}
\usepackage{subfigure}
\usepackage{multirow}
\usepackage{svg}
\usepackage{makecell}
\usepackage{amsmath}
\usepackage{cleveref}
\usepackage{tabularray}
\definecolor{MineShaft}{rgb}{0.2,0.2,0.2}

\usepackage{tabularx}
\usepackage{enumitem}

\title{Generate-on-Graph: Treat LLM as both Agent and KG for \\ Incomplete Knowledge Graph Question Answering}

\author{
Yao Xu\textsuperscript{1,2}, 
\textbf{Shizhu He}\textsuperscript{1,2\thanks{ \; Corresponding Author}}\;,
Jiabei Chen\textsuperscript{1,2},
Zihao Wang\textsuperscript{3},
\textbf{Yangqiu Song}\textsuperscript{3}, \\
\textbf{Hanghang Tong}\textsuperscript{4}, 
\textbf{Guang Liu}\textsuperscript{5},
\textbf{Jun Zhao}\textsuperscript{1,2},
\textbf{Kang Liu}\textsuperscript{1,2} \\
\textsuperscript{1} The Key Laboratory of Cognition and Decision Intelligence for Complex Systems, \\ Institute of Automation, Chinese Academy of Sciences\\ 
\textsuperscript{2} School of Artificial Intelligence, University of Chinese Academy of Sciences \\ 
\textsuperscript{3} The Hong Kong University of Science and Technology \\
\textsuperscript{4} University of Illinois Urbana-Champaign \\
\textsuperscript{5} Beijing Academy of Artificial Intelligence\\ 
\{yao.xu, jzhao, shizhu.he, kliu\}@nlpr.ia.ac.cn, chenjiabei2024@ia.ac.cn
}

\begin{document}
\maketitle
\begin{abstract}
To address the issues of insufficient knowledge and hallucination in Large Language Models (LLMs), numerous studies have explored integrating LLMs with Knowledge Graphs (KGs). However, these methods are typically evaluated on conventional Knowledge Graph Question Answering (KGQA) with complete KGs, where all factual triples required for each question are entirely covered by the given KG. In such cases, LLMs primarily act as an agent to find answer entities within the KG, rather than effectively integrating the internal knowledge of LLMs and external knowledge sources such as KGs. 
In fact, KGs are often incomplete to cover all the knowledge required to answer questions. To simulate these real-world scenarios  and evaluate the ability of LLMs to integrate internal and external knowledge, we propose leveraging LLMs for QA under Incomplete Knowledge Graph (IKGQA), where the provided KG lacks some of the factual triples for each question, and construct corresponding datasets.
To handle IKGQA, we propose a training-free method called Generate-on-Graph (GoG), which can generate new factual triples while exploring KGs. Specifically, GoG performs reasoning through a Thinking-Searching-Generating framework, which treats LLM as both Agent and KG in IKGQA.
Experimental results on two datasets demonstrate that our GoG outperforms all previous methods.

\end{abstract}

\section{Introduction}

\begin{figure}[t]
  \centering
  \includegraphics[width=0.45\textwidth]{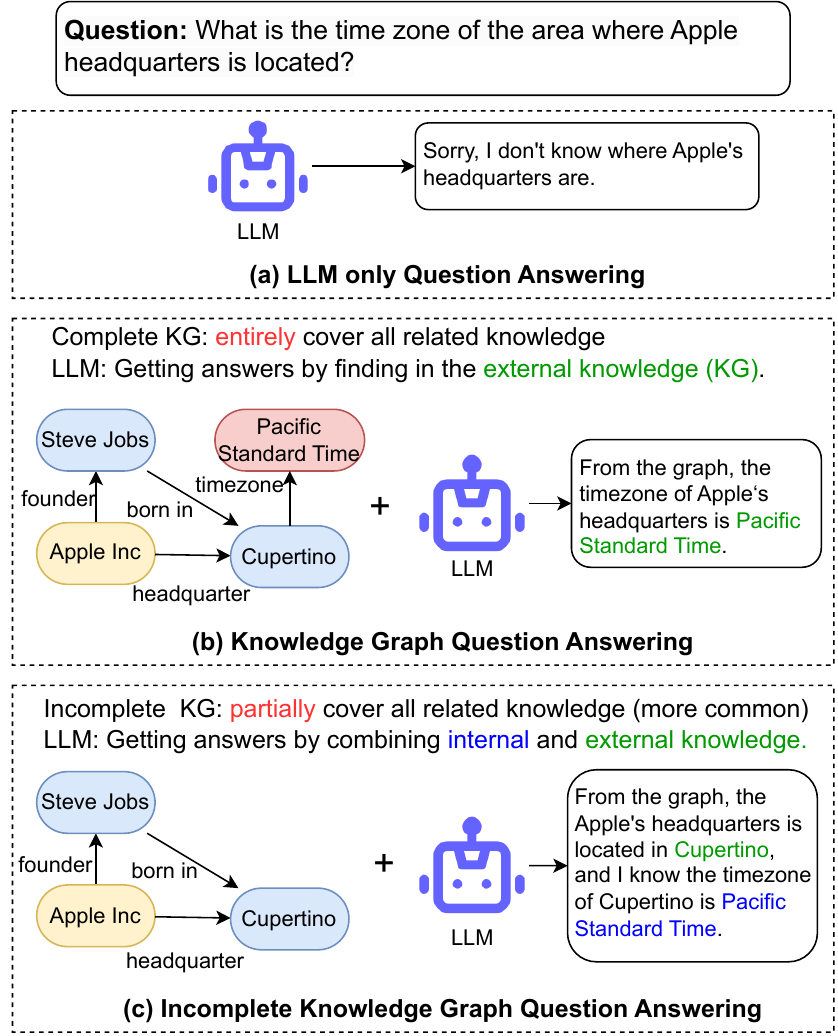} 
  \caption{Comparison between three Question Answering tasks: (a) LLM only QA, (b) Knowledge Graph QA (KGQA), (c) Incomplete Knowledge Graph QA (IKGQA), where the triple \textit{(Cupertino, timezone, Pacific Standard Time)} is missing. The yellow and red nodes represent topic and answer entity, respectively.}
  \label{ckg_ikg}
\end{figure}

Large Language Models (LLMs) \cite{few_shot_learner, eval_chatgpt} have made great success in various natural language processing (NLP) tasks. Benefiting from extensive model parameters and vast amounts of pre-training corpus, LLMs can solve complex reasoning tasks through prompting engineer and in-context learning \cite{icl_survey}, without fine-tuning for specific tasks. 

\begin{figure}[t]
  \centering
  \includegraphics[width=0.45\textwidth]{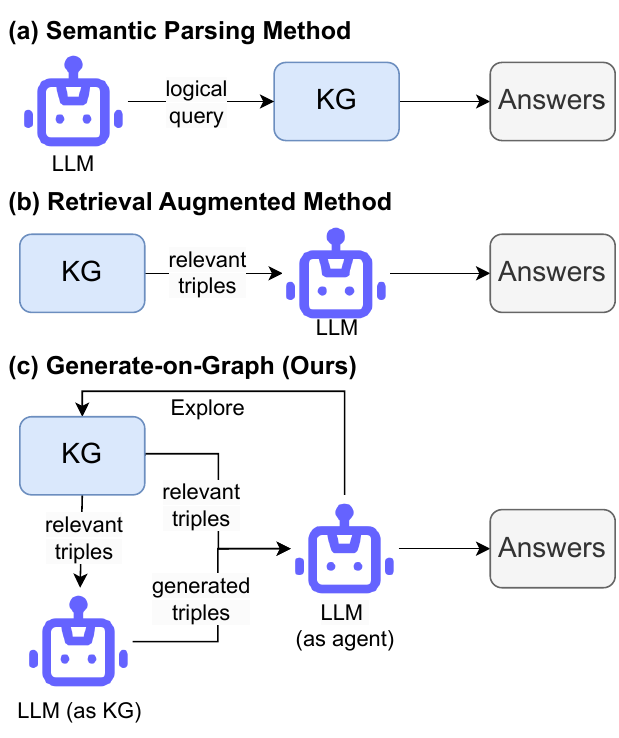} 
  \caption{Three paradigms for combining LLMs with KGs.}
  \label{method_compare}
\end{figure}

However, LLMs still suffer from insufficient knowledge and hallucination issues ~\cite{halu_survey, halu_eval}, as shown in Figure \ref{ckg_ikg} (a). To mitigate those issues, many methods that incorporate LLM with Knowledge Graphs (KGs) \cite{kg_survey_2021} have been proposed \cite{unifying_kg_2023}, where KGs provide accurate factual knowledge in triple format, while LLMs provide strong language processing and knowledge integration ability. These works can be roughly divided into two categories, as shown in Figure \ref{method_compare}: 
(1) \textbf{Semantic Parsing (SP) methods } \cite{kb-binder, luo2023chatkbqa}, which use LLMs to convert natural language questions to logical queries, and then obtain answers by executing these logical queries on KGs. 
(2) \textbf{Retrieval Augmented (RA) methods} \cite{liChainofKnowledgeGroundingLarge2023}, which retrieve information related to the question from KGs as external knowledge to guide LLMs to generate the answers. 

Semantic parsing methods exclusively treat LLMs as parser, which depend heavily on KGs' quality and completeness \cite{tog_2023}. Although retrieval augmented methods claim to solve the drawbacks of semantic parsing methods and obtain good performance on conventional Knowledge Graph Question Answering (KGQA) \cite{kgqa}, it is still hard to verify whether they really integrate knowledge from KGs and LLMs.
One crucial reason is that, \textbf{in conventional KGQA tasks, the factual triples required for each question are entirely covered by the KG}. 
For example, for the question \textit{"What is the timezone of the area where Apple headquarters is located?"} in Figure \ref{ckg_ikg} (b), the LLMs only need to start from \textit{"Apple headquarters"}, sequentially choose the relation predicates \textit{"located\_in"} and \textit{"timezone"} to find the answer. That means, in this scenario, LLMs only need to ground the relationship mentioned in the question to the specific relation predicates in the KG to reach the answer entity \textit{"Pacific Standard Time"} without really integrating internal and external knowledge.

However, on the one hand, KGs are often incomplete to cover all the knowledge required to answer questions in real-world scenarios. For example, for the same question in Figure \ref{ckg_ikg} (c), the crucial triple (\textit{Cupertino, timezone, Pacific Standard Time}) does not exist in the KG. On the other hand, LLMs contain rich knowledge content and possess powerful reasoning ability.  For example, LLMs usually know the time zone of a city. This raises the research question: \textbf{Can LLMs be combined with incomplete KGs to answer complex questions?}

To answer this question, in this paper, we first propose a new benchmark, which utilizes LLMs for QA under incomplete KG (IKGQA), to simulate realistic scenarios. We construct the IKGQA datasets based on existing public KGQA datasets and simulate KGs with varying degrees of incompleteness by randomly dropping triples according to different probabilities. Unlike conventional KGQA, the corresponding KG in IKGQA does not encompass all the factual triplets required for each question.
This means that semantic parsing methods may fail to retrieve the final answer even generating the correct SPARQL query \footnote{Semantic parsing methods always parse "timezone" into to \textit{"timezone"} rather than \textit{"located\_in -> timezone"} because of the training set, more details can be found in Appendix \ref{app:sp_methods}.}. 
Besides, previous retrieval augmented methods also can't perform well under incomplete KGs, as they still heavily rely on the retrieved paths, more details are in Appendix \ref{app:ra_methods}. 
Compared to KGQA, IKGQA holds greater research significance for the following reasons: 
(1) It is closer to real-world scenarios where the given KG is incomplete to answer users' questions.
(2) It can better evaluate the ability of LLMs to integrate the internal and external knowledge.


We also propose a novel method called Generate-on-Graph (GoG) for IKGQA, as illustrated in Figure \ref{method_compare} (c), which not only treats LLM as an agent exploring the given KGs to retrieve relevant triples, but also as a KG to generate additional factual triples for answering this question. Specifically, GoG adopts a Thinking-Searching-Generating framework, consisting of three main steps: 
(1) \textbf{Thinking}: LLMs decompose the question and determine whether to conduct further searches or generate relevant triples based on the current state.
(2) \textbf{Searching}: LLMs use pre-defined tools, such as a KG engineer executing SPARQL queries, to explore the KGs and filter out irrelevant triples.
(3) \textbf{Generating}: LLMs use its internal knowledge and reasoning abilities to generate required new factual triples based on the explored subgraph and verify them.
GoG will repeat these steps until obtaining adequate information to answer the question.
The codes and data are available at \url{https://github.com/YaooXu/GoG}. 

The main contributions of this paper can be summarized as follows:

\begin{enumerate}
  \item We propose leveraging LLMs for QA under incomplete KG (IKGQA) to better evaluate LLMs' ability, and construct corresponding IKGQA datasets based on existing KGQA datasets.
  \item We propose Generate-on-Graph (GoG), which uses the Thinking-Searching-Generating framework, to address IKGQA.
  \item Experimental results on two datasets show the superiority of GoG, and demonstrate
  that LLMs can be combined with incomplete KGs to answer complex questions.
\end{enumerate}

\begin{figure*}[t]
  \centering
  \includegraphics[width=0.98\textwidth]{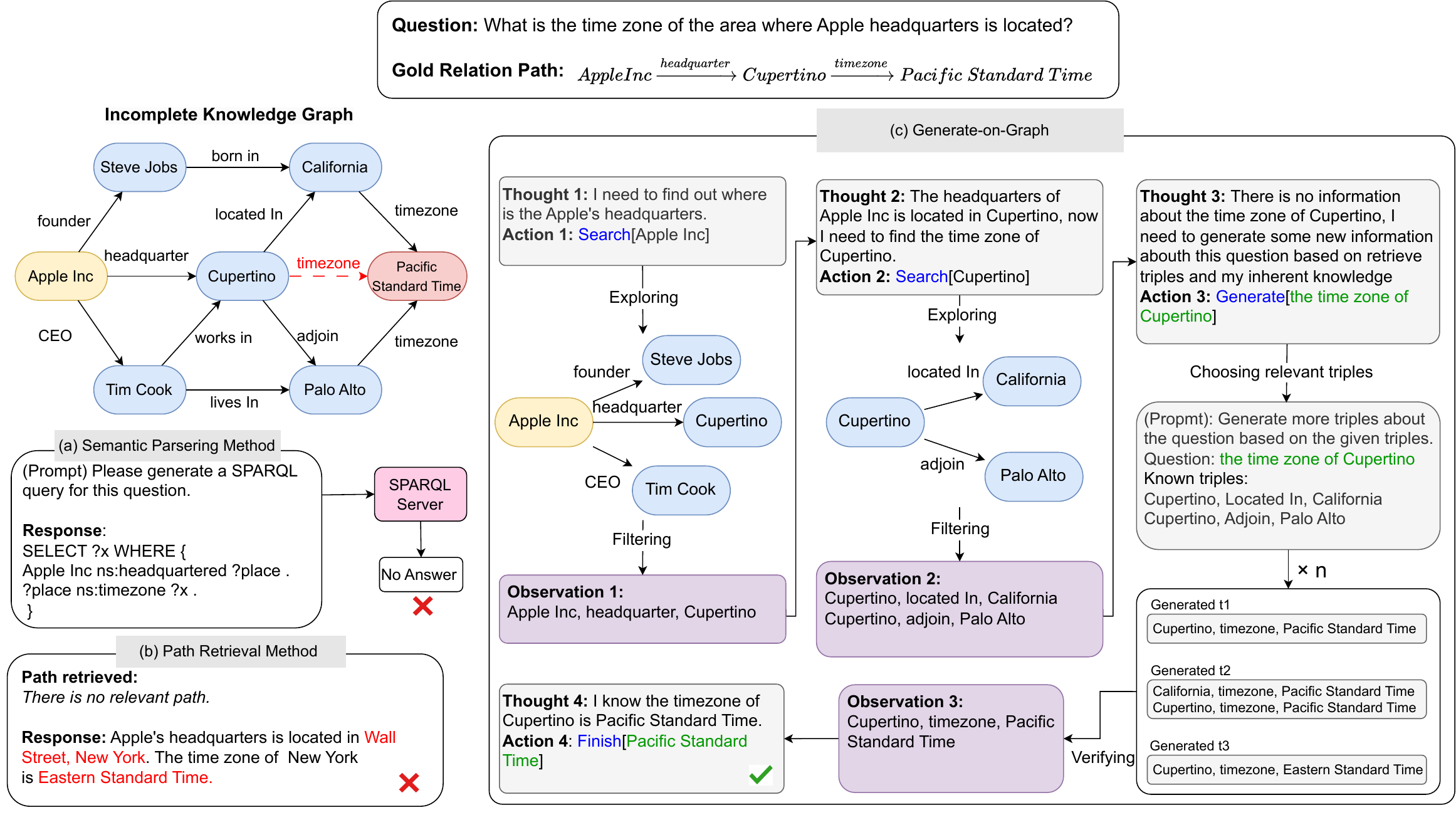} 
  \caption{Comparison of three methods in solving IKGQA: (a) Semantic parsing based method (e.g., ChatKBQA~\cite{luo2023chatkbqa}), (b) Path retrieval method (e.g., ToG~\cite{tog_2023}), (c) The proposed GoG with Thinking-Searching-Generating framework.
  }
  \label{framework}
\end{figure*}

\section{Related Work}

\noindent \textbf{Question Answering under Incomplete KG}. Some previous works \cite{saxena2020improving, zan2022complex, zhao2022improving, guo2023knowledge} attempt to train KG embeddings to predict answers by similarity scores under incomplete KG. Compared to these previous KGE-based works, we propose leveraging LLMs for QA under incomplete KG to study whether LLMs can integrate internal and external knowledge well.

\noindent \textbf{Unifying KGs and LLMs for KGQA}.
Various methods have been proposed to unify KGs and LLMs to solve KGQA, these methods can be classified into two categories: Semantic Parsing (SP) methods and Retrieval Augmented (RA) methods. 
SP methods transform the question into a structural query using LLMs. These queries can then be executed by a KG engine to derive answers based on KGs. These methods generate the drafts as preliminary logical forms first, and then bind the drafts to the executable ones with entity and relation binders, such as  KB-BINDER \cite{kb-binder} and ChatKBQA \cite{luo2023chatkbqa}. However, the effectiveness of these methods relies heavily on the quality of the generated queries and the completeness of KGs. 
RA methods retrieve related information from KGs to improve the reasoning performance \cite{liGraphReasoningQuestion2023b}. 
ToG \cite{tog_2023} treats the LLM as an agent to interactively explore relation paths step-by-step on KGs and perform reasoning based on the retrieved paths. 
RoG \cite{rog_2023} first generates relation paths as faithful plans, and then use them to retrieve valid reasoning paths from the KGs for LLMs to reason.
Readi \cite{cheng_call_2024} generates a reasoning path and edit the path only when necessary.
\citeauthor{salnikov2023large} propose "generate-then-select" method that first uses LLMs to generate answers directly, and then constructs subgraphs and selects the subgraph most likely to contain the correct answer.

Our GoG belongs to retrieval augmented methods, we also utilize the knowledge modeling ability of LLMs, which is also similar to GAG \cite{gag}.

\noindent \textbf{LLM reasoning with Prompting}.
Many works have been proposed to elicit the reasoning ability of LLMs to solve complex tasks through prompting \cite{cot, DecomposedPrompting2023}. 
Complex CoT \cite{fuComplexityBasedPromptingMultiStep2023} creates and refines rationale examples with more reasoning steps to elicit better reasoning in LLMs. 
Self-Consistency \cite{wangSelfConsistencyImprovesChain2023} fully explores various ways of reasoning to improve their performance on reasoning tasks.
DecomP \cite{DecomposedPrompting2023} solves complex tasks by instead decomposing them into simpler sub-tasks and delegating these to sub-task specific LLMs.
ReAct \cite{yaoReActSynergizingReasoning2023} treats LLMs as agents that interact with the environment and make decisions to retrieve information from external source.
GoG can be viewed as a fusion of ReAct and DecomP, thereby enabling a more comprehensive utilization of the diverse capabilities internal in LLMs for addressing complex questions.

\section{Preliminary}
In this section, we first introduce Knowledge Graphs (KGs). Then, we use symbols of KGs to describe relation path and Knowledge Graph Question Answering (KGQA).

\noindent \textbf{Knowledge Graphs (KG)} can be described as a set of inter-linked factual triples, i.e., $\mathcal{G} = \{(h,r,t) \in \mathcal{V} \times \mathcal{R} \times \mathcal{V} \}$, where $h, r \in \mathcal{V}$ denote the head and tail entity,  $r \in \mathcal{R}$ represents the relation.

\noindent \textbf{Knowledge Graph Question Answering (KGQA)} is a reasoning task that aims to predict answer entities $e_a \in \mathcal{A}_q$ based on $\mathcal{G}$. Following previous work \cite{sun2019pullnet}, we call the entities mentioned in question $q$ as topic entities, denoted as $e_t \in \mathcal{T}_q$. Many datasets \cite{cwq, webqsp} give the standard SPARQL query of each question, which demonstrates a relation path from the topic entity $e_t$ to answer entity $e_a$. We call this path as gold relation path, denote it as $w_g = e_q \xrightarrow{r_1} e_1 \xrightarrow{r_2} ... \xrightarrow{r_l} e_a$. For example, the gold relation path of the question in Figure \ref{framework} is $w_g = Apple \; Inc \xrightarrow{headquarter} Cupertino \xrightarrow{timezone} Pacific \; Standard \; Time$. 
In KGQA, $\forall i \in [1, l], \; (e_{i-1}, r_i, e_{i}) \in \mathcal{G}$. That is, it is guaranteed that all triples in gold path are contained by $\mathcal{G}$.

\section{Incomplete Knowledge Graph Question Answering (IKGQA)}

\subsection{Task Introduction}
IKGQA differs from KGQA in that, in IKGQA, $\exists i \in [1, l], \; (e_{i-1}, r_i, e_{i}) \notin \mathcal{G}$. That is, it doesn't guarantee that all triples in gold path are contained by $\mathcal{G}$. For example, the triple \textit{(Cupertino, timezone, Pacific Standard Time)} in $w_g$ may not be contained by $\mathcal{G}$. Therefore, models need to recall them from LLMs or reasoning from subgraph information.

\subsection{Datasets Construction}
\label{construct_dataset}

At present, there are no IKGQA datasets readily available. In this paper, to promote relevant research, we construct two IKGQA datasets based on two widely used KGQA datasets: WebQuestionSP (WebQSP) \cite{webqsp} and Complex WebQuestion (CWQ) \cite{cwq}. Both datasets use Freebase \cite{bollacker2008freebase} as their background KG. 
To simulate incomplete KGs, we randomly delete some crucial triples, which appear in the gold relation path, for each question from the original KG. By doing this, simple semantic parsing methods almost fail to obtain the correct answers. In order to save computational costs, we randomly select 1,000 samples of these two datasets for constructing IKGQA questions.

The process of generating crucial triples of a question is illustrated in Algorithm \ref{gen_ct}.

\begin{algorithm}[]
\SetAlgoLined
\KwIn{SPARQL query $s_q$, KG $\mathcal{G}$, probability $p$}
\KwOut{Dropped crucial triples list $L$}

Initialize $L \leftarrow []$, $filtered\_triples \leftarrow []$\;

$binding\_results$ $\leftarrow$ execute($s_q$, $\mathcal{G}$)\;

$all\_triples$ $\leftarrow$ convert($binding\_results$)\;

// Filter property node (e.g., height, text)

$filtered\_triples$ $\leftarrow$ filter($all\_triples$)\;

\For{each $t$ in $filtered\_triples$}{
    $r$ $\leftarrow$ generate\_random\_float()\;
    \If{$r \leq p$}{
      $L$.add($t$)
    }
}
Return  $L$\;

\caption{Obtaining crucial triples of the question $q$}
\label{gen_ct}
\end{algorithm}



\section{Generate-on-Graph (GoG)}

In this section, we introduce our method Generate-on-Graph (GoG), which can integrate the knowledge of KGs and LLMs, as well as utilize the reasoning ability of LLMs. The workflow of GoG is illustrated in Figure \ref{framework} (c).
GoG utilizes the Thinking-Searching-Generating framework, which consists of three main steps: Thinking, Searching and Generating.

\subsection{Thinking}
Motivated by ReAct \cite{yaoReActSynergizingReasoning2023}, we consider the LLM as an agent interacting with an environment to solve tasks. GoG use the Thinking-Searching-Generating framework to answer questions. As shown in Figure \ref{framework} (c), for each step $i$, GoG first generates a thought $t_i \in \mathcal{L}$, where $\mathcal{L}$ is the language space, to decompose the original question (Thought 1), decide which next sub-question should be solved (Thought 2) or determine whether it has adequate information to output the final answers (Thought 4). 
Then, based on the thought $t_i$, GoG generates an action $a_i \in \mathcal{A}$, where $\mathcal{A}$ is the action space, to search information from the KG (Action 1, 2) or generate more information by reasoning and internal knowledge (Action 3). 


\begin{table*}[t]
\centering

\begin{tabularx}{0.95\textwidth}{l|>{\centering\arraybackslash}X>{\centering\arraybackslash}X|>{\centering\arraybackslash}X>{\centering\arraybackslash}X} 
\hline
\textbf{Method}                 & \multicolumn{2}{c|}{\textbf{CWQ}} & \multicolumn{2}{c}{\textbf{WebQSP}}  \\ 
\hline
\multicolumn{5}{c}{w.o. Knowledge Graph}                                                                   \\ 
\hline
IO prompt                       & \multicolumn{2}{c|}{37.6}         & \multicolumn{2}{c}{63.3}             \\
CoT                             & \multicolumn{2}{c|}{38.8}         & \multicolumn{2}{c}{62.2}             \\
CoT+SC                          & \multicolumn{2}{c|}{45.4}         & \multicolumn{2}{c}{61.1}             \\ 
\hline
                                & CKG           & IKG               & CKG           & IKG                  \\ 
\hline
\multicolumn{5}{c}{w.t. Knowledge Graph /~Fine-tuned}                                                      \\ 
\hline
RoG \cite{rog_2023}             & 66.1          & 54.2              & \textbf{88.6} & 78.2                 \\
ChatKBQA \cite{luo2023chatkbqa} & \textbf{76.5} & 39.3              & 78.1          & 49.5                 \\ 
\hline
\multicolumn{5}{c}{w.t. Knowledge Graph / Not-Training (GPT-3.5)}                                          \\ 
\hline
KB-BINDER \cite{kb-binder}      & -             & -                 & 50.7          & 38.4                 \\
StructGPT \cite{structgpt_2023} & -             & -                 & 76.4          & 60.1                 \\
ToG \cite{tog_2023}             & 47.2          & 37.9              & 76.9          & 63.4                 \\
GoG (Ours)                      & 55.7          & 44.3              & 78.7          & 66.6                 \\ 
\hline
\multicolumn{5}{c}{w.t. Knowledge Graph / Not-Training (GPT-4)}                                            \\ 
\hline
ToG \cite{tog_2023}             & 71.0          & 56.1              & 80.3          & 71.8                 \\
GoG~(Ours)                      & 75.2          & \textbf{60.4}     & 84.4          & \textbf{80.3}        \\
\hline
\end{tabularx}
\caption{The Hits@1 scores of different models over two datasets under different settings  (\%). CKG and IKG denote using 
complete and incomplete KG (IKG-40\%), respectively. Results of the other baselines were re-run by us \footnotemark[2]. The boldface indicates the best result.}
\label{main_results}
\end{table*}

\subsection{Searching}
The search action is invoked by GoG in the form of $Search[e_i]$, where $e_i$ is the target entity, as illustrated in Action 1 and 2 in Figure \ref{framework} (c). While it is possible to search multiple target entities, like $Search[e_i^1, e_i^2, \ldots]$, for simplicity, we only consider searching for one target entity here. This action aims to find the most relevant top-k entities $E_i$ from the neighboring entities of the target entity $e_i$ based on the last thought $t_i$. The search action consists of two steps: Exploring and Filtering.
\begin{itemize}
    \item \textbf{Exploring} GoG first uses predefined SPARQL queries to obtain all the relations $R_i$ linked to the target entity $e_i$. For example, in Figure \ref{framework} (c), $e_1$=\{\textit{Apple Inc}\} $R_1$=\{\textit{founder, headquarter, CEO}\}.
    \item \textbf{Filtering} After retrieving the relation set $R_i$, LLMs are utilized to select the most relevant top-N relations $R'_i$ based on the last thought $t_i$. The prompt used for this step is detailed in Appendix \ref{prompts}. In the case of Figure \ref{framework} (c),  LLMs select $R'_1$=\{\textit{headquarter}\} from $R_1$=\{\textit{founder, headquarter, CEO}\} to answer the thought $t_1$ \textit{"I need to find out where is the Apple's headquarters"}.
\end{itemize}

Finally, we obtain the most relevant entity set $E_i$ based on the target entity $e_t$ and the relevant relation set $R'_i$. As shown in Figure \ref{framework} (c), the Observation in step one is \{(\textit{Apple Inc, headquarter, Cupertino})\}, which is attached to the context to enable GoG to generate the next thought.


\subsection{Generating}
When there is no direct answer from previous Observation, the Generate Action is invoked by GoG in the form of $Generate[t_i]$, where $t_i$ is the last thought, as illustrated in Action 3 in Figure \ref{framework} (c). This action tries to utilize the LLM to generate new factual triples based on retrieval information and internal knowledge. There are three steps in each Generate Action: choosing, generating and verifying.

\begin{itemize}
    \item \textbf{Choosing} To provide LLMs some relevant information to generate more accurate triples, we use BM25 \cite{BM25} to retrieve the most relevant triples from previous Observation. For example, in the Action 3 in Figure \ref{framework} (c), we choose \{(\textit{Cupertino, located\_in, California}), (\textit{Cupertino, adjoin, Palo Alto})\} from Observation 1 and 2 as the relevant triples used in LLM generating new triples.
    \item \textbf{Generating} After retrieving relevant triples, LLMs are utilized to generate new factual triples based on these relevant triples and their internal knowledge. The generating process will be repeated $n$ times to minimize error and hallucination. As shown in Action 3 of Figure \ref{framework} (c), given relevant triples, LLMs generate \{(\textit{Cupertino, timezone, Pacific Standard Time})\} in generated $t_1$. 
    \item \textbf{Verifying} In the end, we use LLMs to verify the generated triples and choose those are more likely to be accurate as the Observation, the prompt used here is shown in Appendix \ref{prompts}. As shown in Observation 3 of Figure \ref{framework} (c), LLMs only remain \{(\textit{Cupertino, timezone, Pacific Standard Time})\} from all generated triples.
\end{itemize}
It is also possible for the LLMs to generate an entity that is not explored before. Therefore, we have to link the entity to its corresponding Machine Identifier (MID) in the KG. This entity linking process is divided into two steps: (1) We retrieve some similar entities and their corresponding types based BM25 scores. (2) We utilize the LLM to select the most relevant entity based on the types, the prompt we use is demonstrated in Appendix \ref{prompts}.

\noindent GoG repeats the above three steps until it obtains adequate information, and then outputs the final answer in the form of $Finish[e_a]$, where $e_a$ represents the answer entity. It should be noticed that the agent could also generate "$Finish[unknown]"$, which means that there is not enough information for the agent to answer the question. In this case, we would roll back and search one more hop neighbors of the last target entity.

\begin{table*}[t]
\centering

\begin{tabularx}{0.95\textwidth}{l*{5}{>{\centering\arraybackslash}X}} 
\hline
\multicolumn{1}{c}{\multirow{2}{*}{\textbf{Method}}} & \multicolumn{5}{c}{\textbf{CWQ}}   
                 \\ 
\cline{2-6}
                                                     & CKG           & IKG-20\%      & IKG-40\%      & IKG-60\%      & IKG-80\%       \\ 
\hline
ToG                                                  & 47.2          & 40.5          & 37.9          & 33.7          & 31.4           \\
GoG                                                  & \textbf{55.7} & \textbf{44.9} & \textbf{44.3} & \textbf{36.2} & \textbf{34.4}  \\
\hline
                                                     & \multicolumn{5}{c}{\textbf{WebQSP}}  
                                                     \\ 
\cline{2-6}
\multicolumn{1}{c}{}                                 & CKG           & IKG-20\%      & IKG-40\%      & IKG-60\%      & IKG-80\%       \\ 
\cline{1-6}
StructGPT                                            & 76.0          & 67.8          & 60.1          & 51.7          & 43.7           \\
ToG                                                  & 76.9          & 70.3          & 61.4          & 60.6          & 55.9           \\
GoG                                                  & \textbf{78.7} & \textbf{70.8} & \textbf{66.6} & \textbf{62.6} & \textbf{56.5}  \\ 
\hline
\end{tabularx}

\caption{The Hits@1 scores of prompt based methods (w/ GPT-3.5) under different numbers of missing triples (\%). CKG represents using the complete KG. IKG-20\%/40\%/60\%/80\% represent randomly drop 20\%/40\%/60\%/80\% crucial triples for each question.}
\label{tab:num_triples}

\end{table*}

\begin{table}[]
\centering
\resizebox{0.45\textwidth}{!}{%
\begin{tabular}{lccc}
\hline
\multicolumn{1}{c}{\multirow{2}{*}{Method}} & \multicolumn{3}{c}{\textbf{CWQ}}                                                  \\ 
\cline{2-4}
\multicolumn{1}{c}{}                        & CKG                     & IKG-40\%                     & NKG                      \\ 
\hline
GoG w/GPT-3.5                               & 55.7                    & 44.3                         & 38.8                     \\
GoG w/Qwen-1.5                              & 63.3                    & 49.2                         & 47.0                     \\
GoG w/Llama-3                                & 59.6                    & 54.6                         & 54.0                     \\
GoG w/GPT-4                                 & \textbf{75.2}           & \textbf{60.4}                & \textbf{55.6}            \\ 
\hline
                                            & \multicolumn{3}{c}{\textbf{WebQSP}}                                               \\ 
\cline{2-4}
                                            & \multicolumn{1}{l}{CKG} & \multicolumn{1}{l}{IKG-40\%} & \multicolumn{1}{l}{NKG}  \\ 
\hline
GoG w/GPT-3.5                               & 78.7                    & 66.6                         & 62.6                     \\
GoG w/Qwen-1.5                              & 77.9                    & 70.2                         & 65.1                     \\
GoG w/Llama-3                                & 77.4                    & 74.4                         & 70.8            \\
GoG w/GPT-4                                 & \textbf{84.4}           & \textbf{80.3}                & \textbf{75.7}                    \\
\hline
\end{tabular}
}
\caption{The Hits@1 scores of GoG using different backbone models (\%). CKG, IKG-40\% and NKG denote using complete, incomplete and no KG. Qwen-1.5 and Llama-3 represent Qwen-1.5-72b-chat and Llama-3-70b-Instruct, respectively.}
\label{different_llms}
\end{table}

\label{Experiments}

\section{Experiments}

\footnotetext[2]{The evaluation strategy we use differs from that of ToG, which makes the performance of ToG vary from those reported. Further details are available in Appendix \ref{app:baselines}.}

\subsection{Experiments Setup}

\paragraph{Evaluation Metrics}
Following previous works \cite{liChainofKnowledgeGroundingLarge2023, structgpt_2023, tog_2023}, we use Hits@1 as our evaluation metric, which measures the proportion of questions whose top-1 predicted answer is correct.

\paragraph{Baselines}
The baselines we compare can be divided into three groups: 
(1) LLM only methods, including standard prompting (IO prompt) \cite{few_shot_learner}, Chain-of-Thought (CoT) prompting \cite{cot} and Self-Consistency (SC) \cite{wangSelfConsistencyImprovesChain2023}. 
(2) Semantic Parsing (SP) methods, including KB-BINDER \cite{kb-binder} and
ChatKBQA \cite{luo2023chatkbqa}.
(3) Retrieval Augmented (RA) methods, including StructGPT \cite{structgpt_2023}, RoG \cite{rog_2023} and ToG \cite{tog_2023}, where RoG is the SOTA among all models requiring fine-tuning.

\paragraph{Experiment Details} 
We use four LLMs as the backbone in our experiments: GPT-3.5, GPT-4, Qwen-1.5-72B-Chat \cite{bai2023qwen} and LLaMA-3-70B-Instruct \cite{touvron2023llama2}.
We use OpenAI API to call GPT-3.5 and GPT-4 \footnotemark[3]. The maximum token length for each generation is set to 256. The temperature parameter is set to 0.7. We use 3 shots in GoG prompts for all the datasets. The prompts we use are listed in Appendix \ref{prompts}. 
\footnotetext[3]{The specific versions of GPT-3.5 and GPT-4 are gpt-3.5-turbo-0613 and gpt-4-0613.}

\paragraph{Datasets Details} For each dataset, we generate four incomplete KGs with varying degrees of completeness: IKG-20\%/40\%/60\%/80\%, representing randomly drop 20\%/40\%/60\%/80\% crucial triples for each question. In addition to the crucial triples themselves, all relations between these two entities will also be deleted. The statistics of these IKGs can be found in Appendix \ref{app:IKG_topic_entity}. 

\subsection{Main Results}

Table \ref{main_results} shows the Hits@1 scores of GoG and all baselines on two datasets under different settings. 
From the table, we can find that, compared with other prompt based methods, GoG can achieve the state-of-the-art performance on CWQ and WebQSP under both complete and incomplete KG settings. 

Under the CKG setting, the main reasons our GoG outperforms ToG are: (1) GoG decompose the problem into sub-problems each step and focuses on the information needed for each sub-problem during the search process, whereas ToG lacks overall planning, making it prone to repetitive exploration or getting lost during the search. (2) GoG adopts a dynamic subgraph expansion search strategy, while ToG only explores some paths. Therefore, the relevant information obtained in GoG is richer. Moreover, this strategy can better handle compound value types (CVTs), as detailed in Appendix \ref{app:cvt}. A case study is 
shown in Appendix \ref{app:cmp_ckg}.

Under the IKG setting, the performance of SP methods significantly declines. This is expected, as these SP methods don't interact with the KGs, which means they have no idea of the absence of some triples.
The performance of ToG and StructGPT on IKG is even worse than that without KG, indicating that these methods still play a role of finding answers rather than effectively integrating internal and external knowledge sources.
Our GoG mitigates this issue by using the \textbf{Generate} Action, which utilizes the LLM to generate new factual triples when no direct answer is found. A case study illustrating this is provided in Appendix \ref{app:cmp_ikg}, and a detailed analysis of the answers generated by GoG can be found in Appendix  \ref{app:result_analysis}.

\begin{figure*}[t]
  \centering
  \includegraphics[width=0.95\textwidth]{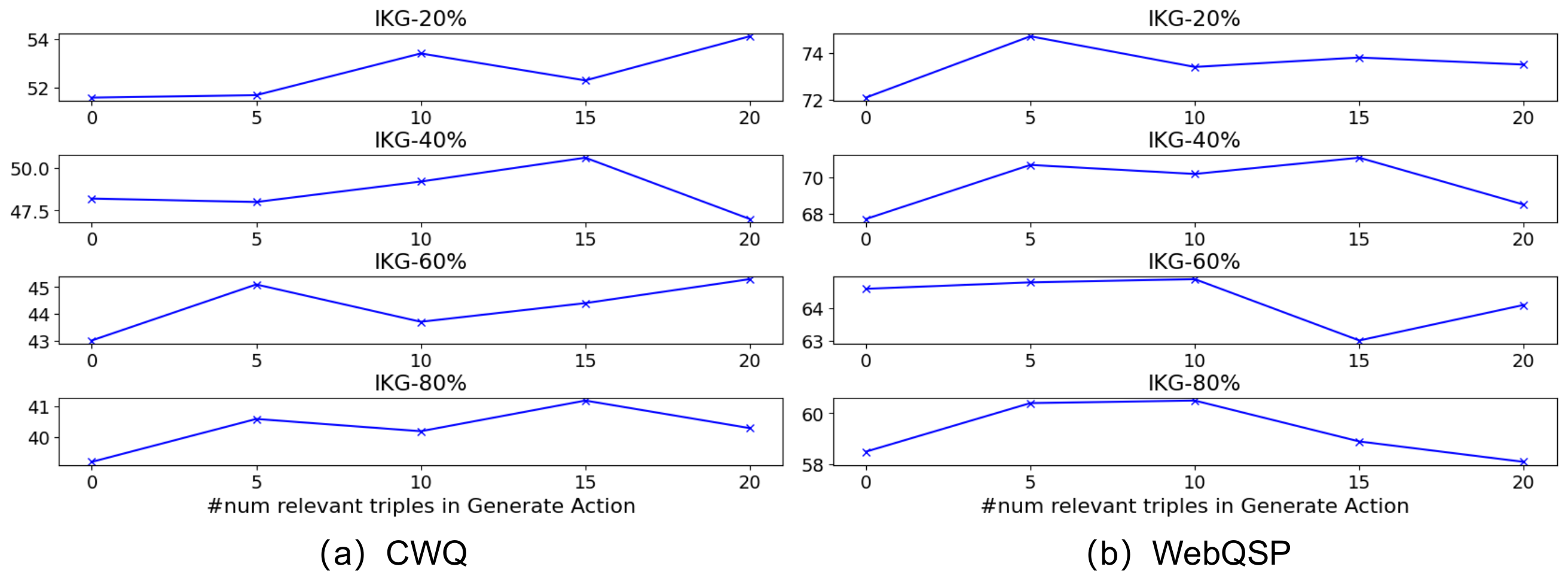}
  \caption{The Hits@1 scores of GoG with different number of related triples in the \textbf{Generate} Action on the CWQ (a) and WebQSP (b) (\%). The backbone LLM is Qwen-1.5-72b-chat.}
  \label{n_related_triples}
\end{figure*}

\subsection{Performance under Different Degrees of KG Incompleteness}
To investigate how different degrees of KG incompleteness affect different methods, we evaluate the performance of methods (w/ GPT-3.5) under KGs with varying degrees of incompleteness, the results are demonstrated in Table \ref{tab:num_triples}.

It can be found that our GoG outperforms other prompt based methods in different degrees of incompleteness. 
Especially on the CWQ dataset, our GoG has a significant improvement on Hits@1 score, achieving average 5.0\% improvement. That emphasizes the importance of integrate the external and internal knowledge of LLMs under incomplete KGs. 
On the contrary, the performance of ToG on IKG-40\% is even lower than that without KG, indicating the performance of ToG still depends heavily on the completeness of KGs.

Even though the majority of questions in the WebQSP dataset are single-hop questions, GoG still outperforms ToG and StructGPT. This is because GoG can leverage the neighboring information of the topic entities to predict the tail entities while other methods can not make full use these information, a case study is shown in Appendix \ref{app:cmp_ikg}.


\subsection{Performance with Different LLMs}


We evaluate how different backbone models affect GoG performance. Table \ref{different_llms} demonstrates
that the performance of GoG using GPT-4 as backbone improves significantly. Especially under complete KGs setting, GoG (w/GPT-4) achieves 84.4 and 75.2 Hits@1 score on the WebQSP and CWQ datasets respectively, which achieve SOTA performance in prompt based methods and outperforms most fine-tuned methods. 

Additionally, we observe that under the NKG setting, Llama-3 consistently outperforms Qwen-1.5, whereas under the CKG setting, the opposite is true. This suggests that the proficiency of LLM as a KG and as an agent is not entirely equivalent. Exploring how different LLMs can leverage their strengths in playing specific roles could be a direction for future research.


\begin{table}
\centering
\resizebox{0.45\textwidth}{!}{%
\begin{tabular}{lcc} 
\hline
\multicolumn{1}{c}{\multirow{2}{*}{Method}} & \multicolumn{2}{c}{\textbf{CWQ }}     \\ 
\cline{2-3}
\multicolumn{1}{c}{}                        & CKG           & IKG-40\%              \\ 
\hline
GoG w.o. Generate                           & 62.7          & 48.6                  \\
GoG w.t. Generate                                         & \textbf{63.3} & \textbf{50.6}         \\ 
\hline
                                            & \multicolumn{2}{c}{\textbf{WebQSP }}  \\ 
\cline{2-3}
                                            & CKG           & IKG-40\%              \\ 
\hline
GoG w.o. Generate                           & 74.7          & 69.4                  \\
GoG w.t. Generate                                         & \textbf{77.9} & \textbf{71.1}         \\
\hline
\end{tabular}
}
\caption{The Hits@1 scores of GoG w.t./w.o. Generate Action (\%).}
\label{generate_action}
\end{table}

\subsection{Ablation Study}

\noindent \textbf{The Effect of the Number of Related Triples}

We perform additional experiments to find out how the number of related triples effect GoG's performance. We select the top-k relevant triples based on BM25, as shown in Figure \ref{n_related_triples}. The results indicate that:
(1) GoG's performance significantly improves with relevant subgraphs, likely because these subgraphs activate LLMs' memory to generate more accurate triples and enable reasoning of new factual triples based on these subgraphs.
(2) In most cases, performance initially increases and then decreases as the number of related triples grows. This decline is mainly due to the introduction of noisy and unrelated knowledge.

\noindent \textbf{The Effect of Generate Action}

We investigate the effect of the Generate Action, as shown in Figure \ref{generate_action}. GoG's performance is lower without Generate Action, confirming the effectiveness of Generate Action. However, GoG without Generate Action still achieves competitive results because it becomes a pure exploring agent, leading to two outcomes: (1) No false negatives, as all answers come from KGs, and (2) It thoroughly searches KGs for answers, whereas GoG with Generate Action may determine to invoke Generate Action instead of continuing the search.

\section{Conclusion}

In this paper, we propose leveraging LLMs for QA under Incomplete KGs (IKGQA), and construct relevant datasets. We propose Generate-on-Graph (GoG), which can effectively integrate the external and internal knowledge of LLMs.
Experiments on two datasets show the superiority of GoG, and demonstrate that an LLMs can be combined with incomplete KGs to answer complex questions.

\section*{Limitation}

The limitations of our proposed GoG are as follows: 
(1) It is possible for LLM to hallucinate in the  
Generate Action, which is unavoidable for existing LLMs. 
(2) There is room for further improvement in performance, as GoG's performance is lower than that with CoT prompt when KGs are very incomplete.

\section*{Ethics Statement}
This paper proposes a method for complex question answering in incomplete knowledge graph, and the experiments are conducted on public available datasets. 
As a result, there is no data privacy concern. Meanwhile, this paper does not involve human annotations, and there are no related ethical concerns.

\section*{Acknowledgment}
This work was supported by Beijing Natural Science Foundation (L243006) and the National Natural Science Foundation of China (No.62376270). This work was supported by the Youth Innovation Promotion Association CAS.

\bibliography{custom}
\bibliographystyle{acl_natbib}

\appendix
\newpage

\begin{table*}
\centering
\begin{tblr}[t]{
  width = 0.9\linewidth,
  colspec = {Q[104]Q[837]},
  cell{1}{2} = {fg=MineShaft},
  cell{2}{1} = {fg=MineShaft},
  vline{2} = {-}{},
  hline{1-3} = {-}{},
}
Question & In the nation that spends the Bahamian dollar as currency, what time zone is used?                                                                                                                                                                                        \\
SPARQL   & {PREFIX ns: <http://rdf.freebase.com/ns/>\\SELECT DISTINCT ?x WHERE \{\\~ FILTER (?x != ?c) FILTER (!isLiteral(?x) OR lang(?x) = '' OR langMatches(lang(?x), 'en'))\\~ ?c ns:location.country.currency\_used ns:m.01l6dm .\\~ ?c ns:location.location.time\_zones ?x .\\\}} 
\end{tblr}
\caption{An example about "timezone" in the CWQ dataset.}
\label{tab:timezone}
\end{table*}

\section{Semantic Parsing Methods Details}
\label{app:sp_methods}
The training datasets for SP methods are constructed under the complete KGs, which means that "Time Zone" corresponds directly to the relation \textit{"ns:location.location.time\_zones"} rather than a two-hop path \textit{"ns:location.located\_in -> ns:location.location.time\_zones"}. An example in CWQ is shown in Table \ref{tab:timezone}. This means SP models trained on CWQ will always output \textit{"?c ns:location.location.time\_zones ?x"} instead of \textit{"?c ns:location.located\_in ?y . ?y ns:location.location.time\_zones ?x"}. Therefore, these methods will fail under Incomplete KGs. In another word, semantic parsing methods don't interact with the KGs, which means they have no idea of the absence of some triples.

\section{Retrieval Augmented Methods Details}
\label{app:ra_methods}

The RA method retrieves relevant paths from the knowledge graph (KG) and uses these paths as context for the large language model (LLM) to generate answers. For instance, ToG employs an LLM to explore the KG, using beam search to select paths related to the question. However, analysis of ToG's results reveals that approximately 70\% of the correct answers come directly from the explored paths, and less than 10\% of the correct answers are derived from a combination of the explored path knowledge and the internal knowledge of the LLM. Subsequent experimental results also indicate that under the IKG setting, ToG's performance is even inferior to that of using the LLM alone. This further demonstrates that such methods do not truly integrate the internal of LLMs and external knowledge of KGs.

\section{Prompt List}
\label{prompts}

The prompts used in GoG are shown in Table \ref{tab:prompts}.

\section{Settings for Baselines}
\label{app:baselines}

Following ToG, the Freebase dump is acquired from  \url{https://developers.google.com/freebase?hl=en}, we deploy Freebase with Virtuoso. GoG, RoG, KB-BINDER and ChatKBQA are evaluated on the same Freebase database.

\noindent \textbf{RoG}. We use the checkpoints and the default settings provided by the official repository: n\_beam=3 in generating rule, max\_new\_tokens=512 in inferring answers.

\noindent \textbf{ChatKBQA}. We use the predicted S-expression provided by the official repository, and convert them into SPARQL queries. To compare ChatKBQA with other models fairly, we execute these SPARQL queries under the Freebase database mention before instead the DB files provided by them. Therefore, the performance of ChatKBQA reported in Table \ref{main_results} is slightly different from that in their original paper.

\noindent \textbf{KB-BINDER}. We use the official repository and use KB-BINDER (6)-R (with majority vote and retrieve the most similar exemplars) to infer answers. However, the code-davinci-002 used in their original paper is not available, so we use GPT-3.5 instead. Besides, to reduce runtime, we decreased the number of candidate MID combinations (despite that, it still takes about 4 hours to answer 200 questions). Therefore, the performance of KB-BINDER reported in Table \ref{main_results} is slightly different from that in their original paper.

\noindent \textbf{ToG}. We use the official repository and their default settings for inferring answers: max\_length=256, width=3, depth=3. Since the official repository doesn't provide the alias answers in the CWQ dataset, we evaluate ToG on the CWQ dataset without considering alias answers (the same strategy for all models). Therefore, the performance of ToG reported in Table \ref{main_results} is slightly different from that in their original paper.

\noindent \textbf{StructGPT}. We use the official repository and running scripts to evaluate StructGPT on the WebQSP dataset.

\section{Statistics of Topic Entities in IKGs}
\label{app:IKG_topic_entity}

The statistics of dropped edges are shown in Table \ref{tab:mean_delete_edges}. Besides, we also ensure that after deleting these crucial triples, the number of neighbor nodes of the topic entities will not be zero. The statistics of topic entities are shown in Table \ref{tab:ikg_topic_entities}, and we drop those samples which have isolated topic entities (topic entity without any neighbor node).

\begin{table}
\centering
\begin{tblr}{
  width = \linewidth,
  colspec = {Q[183]Q[183]Q[183]Q[183]Q[183]},
  column{even} = {c},
  column{3} = {c},
  column{5} = {c},
  hlines,
}
                & \textbf{IKG-20\%} & \textbf{IKG-40\%} & \textbf{IKG-60\%} & \textbf{IKG-80\%} \\
\textbf{CWQ}    & 2.2               & 4.3               & 6.4               & 7.9              \\
\textbf{WebQSP} & 6.6               & 13.9              & 20.3              & 27.4              
\end{tblr}
\caption{The average number of edges deleted for each question under different incompleteness degrees.}
\label{tab:mean_delete_edges}
\end{table}

\begin{table*}
\centering
\begin{tblr}{
  width = 0.9\linewidth,
  colspec = {Q[108]Q[400]Q[80]Q[80]Q[80]Q[80]},
  column{3} = {c},
  column{4} = {c},
  column{5} = {c},
  column{6} = {c},
  cell{2}{1} = {r=2}{},
  cell{2}{2} = {fg=MineShaft},
  cell{4}{1} = {r=2}{},
  cell{4}{2} = {fg=MineShaft},
  hline{1,2,4,6} = {-}{},
}
\textbf{Dataset} &                                                & \textbf{IKG-20\%} & \textbf{IKG-40\%} & \textbf{IKG-60\%} & \textbf{IKG-80\%} \\
CWQ              & Median number of neighbor nodes                & 27             & 26             & 27             & 27             \\
                 & Number of isolated topic entities  & 19             & 42             & 59             & 53             \\
WebQSP           & Median number of neighbor nodes                & 428            & 427            & 427            & 426            \\
                 & Number of isolated topic entities & 1              & 2              & 1              & 2              
\end{tblr}
\caption{Statistics of topic nodes in Incomplete KGs. Isolated topic entity represent topic entity without any neighbor node.}
\label{tab:ikg_topic_entities}
\end{table*}

\section{Compound Value Type (CVT) node}
\label{app:cvt}
\begin{figure}[t]
  \centering
  \includegraphics[width=0.45\textwidth]{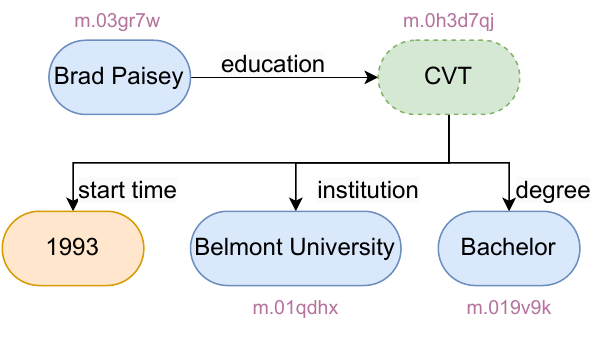} 
  \caption{An example of compound value types (CVTs) in Freebase dataset. Blue, green and orange nodes denote normal entities, CVT node and property node.}
  \label{cvt}
\end{figure}

Compound Value Type (CVT) nodes are usually utilized to model events, which could involve start time, end time, location and so on, in KGs. An example of CVT node is illustrated in Figure \ref{cvt}.

\section{Result Analysis}
\label{app:result_analysis}

\subsection{Performance under Generate Action}
Table \ref{tab:generate_analysis} illustrates the frequency of the Generate operation in different datasets alongside their corresponding Hits@1 scores. In the complete KGs setting, GoG still conducts the Generate operation when related relations are not correctly selected or when answers to sub-questions cannot be directly found via a one-hop relationship. In the incomplete KGs setting, the frequency of the Generate operation is higher, as GoG needs to generate new factual triples that are missing in the KGs.
Hits@1 scores under both settings mean that most generation leading to correct results. 

\subsection{Error Analysis}
\begin{figure*}[t]
  \centering
\includegraphics[width=0.98\textwidth]{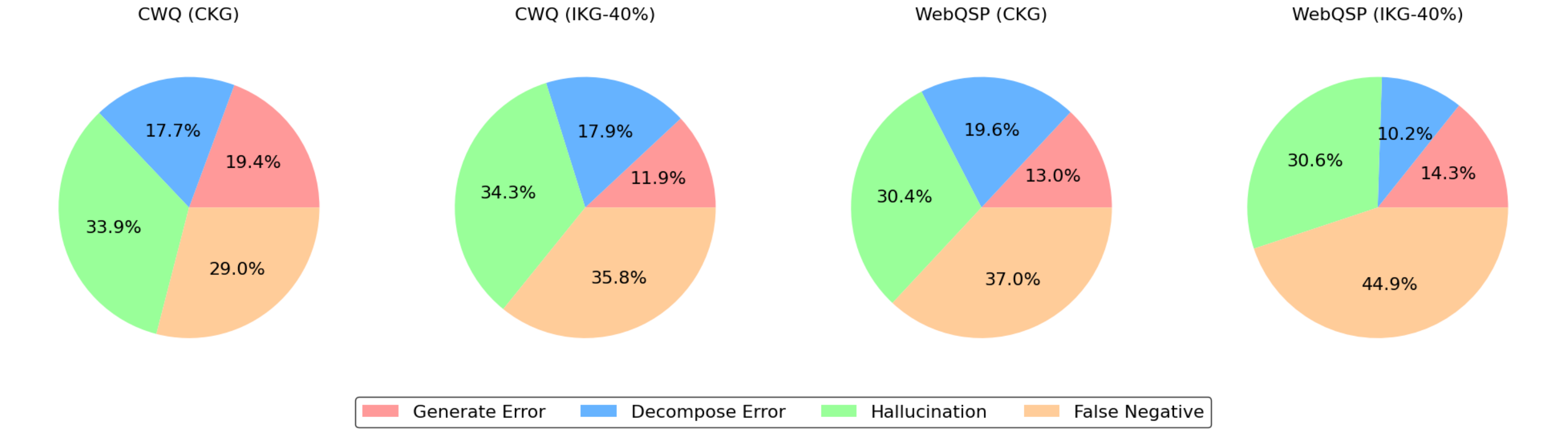} 
  \caption{The error proportions of GoG under different datasets and settings.}
  \label{error_analysis}
\end{figure*}

We consider four types of errors: 
(1) Generate Error, LLMs make error in the Generate Action, such as output wrong entities or "unknown". 
(2) Decompose Error, LLMs forget the original question after multi-round searching and answer the wrong sub-question in the end.
(3) Hallucination, the final answer produced by the LLM is not supported by the evidence in the context (e.g., it lacks some of the constraints), yet the LLM still believes this answer satisfies all the constraints of the question.
(4) False Negative, LLMs output the alias name of the ground truth.
The distribution is shown in Figure \ref{error_analysis}. It is evident that the majority of actual errors stem from hallucinations, discounting false negative samples. Moreover, under the IKG setting, there is a higher likelihood of False Negative occurrences due to discrepancies between the answers generated by the Generate Action and the reference answers (for instance, the LLM outputs 'The US' while the correct answer is "America").

\begin{table*}
\centering

\def\arraystretch{1.5}
\begin{tabular}{lccccc} 
\hline
\multicolumn{1}{c}{\multirow{2}{*}{Models}} & \multicolumn{5}{c}{CWQ}                                                        \\ 
\cline{2-6}
\multicolumn{1}{c}{}                        & CKG           & IKG-20\%      & IKG-40\%      & IKG-60\%      & IKG-80\%       \\ 
\hline
GoG w/GPT-3.5                               & 21.0\% (53.8) & 33.8\% (45.5) & 35.9\% (52.9) & 39.1\% (45.2) & 39.8\% (48.7)  \\
GoG w/Qwen-1.5                              & 24.2\% (44.2) & 35.5\% (42.2) & 40.0\% (43.5) & 46.5\% (41.7) & 50.4\% (43.6)  \\ 
\hline
                                            & \multicolumn{5}{c}{WebQSP}                                                     \\ 
\cline{2-6}
                                            & CKG           & IKG-20\%      & IKG-40\%      & IKG-60\%      & IKG-80\%       \\ 
\hline
GoG w/GPT-3.5                               & 19.3\% (63.2) & 24.4\% (63.5) & 26.8\% (66.4) & 32.5\% (57.8) & 38.2\% (66.4)  \\
GoG w/Qwen-1.5                              & 23.4\% (55.9) & 28.2\% (51.4) & 33.9\% (57.8) & 37.7\% (60.2) & 49.5\% (56.5)  \\
\hline
\end{tabular}

\caption{Ratio of Generate operation in different KG settings. Numbers in brackets represent corresponding Hits@1 score.}
\label{tab:generate_analysis}
\end{table*}

\begin{table*}
\centering
\begin{tabular}{>{\hspace{0pt}}m{0.1\linewidth}>{\hspace{0pt}}m{0.8\linewidth}} 
\hline
\multicolumn{1}{>{\centering\hspace{0pt}}m{0.1\linewidth}}{\textbf{Tasks}} & \multicolumn{1}{>{\centering\arraybackslash\hspace{0pt}}m{0.8\linewidth}}{\textbf{Prompt}}                                                                                                                                                                                                                                                                                                                                                                                                                                                                                                                                                                                                                                                                                                                                                                                                                                                                                                                                                                                                                                                                                                                                                                                                                                                                                                                                                                                                                    \\ 
\hline
\textbf{GoG}\par{}\textbf{Instruction}                                       & Solve a question answering task with interleaving Thought, Action, Observation steps. Thought can reason about the current situation, and Action can be three types:\par{}(1) Search[entity1 \textbar{} entity2 \textbar{} ...], which searches the exact entities on Freebase and returns their one-hop subgraphs. You should extract the all concrete entities appeared in your last thought without redundant words, and you should always select entities from topic entities in the first search.\par{}(2) Generate[thought], which generate some new triples related to your last thought. These new triples may come from your inherent knowledge directly or reasoning from the given triples.\par{}(3) Finish[answer1 \textbar{} answer2 \textbar{} ...], which returns the answer and finishes the task. The answers should be complete entity label appeared in the triples. If you don't know the answer, please output Finish[unknown].\par{}Entities and answers should be separated by "\textbar{}".\par{}Attention please, entities begin with "m." (e.g., m.01041p3) represent CVT (compound value type) node, and they shouldn't be selected as the final answers. To find out those entities involved in these event, you could select them as the entities to be searched. You should generate each step without redundant words.\par{}Here are some examples.\par{}\textit{In-Context Few-shot}\par{}Question: \{Question\}\par{}Topic Entity: \{List of Topic Entities\}\par{}Thought 1:  \\ 
\hline
\textbf{Filter}\par{}\textbf{Relations}                                      & Please select 3 relations that most relevant to the question and rank them. You should answer these relations in list format directly without redundant words.~\par{}Here are some examples.\par{}\textit{In-Context Few-shot}\par{}Thought: \{Thought\}\par{}Entity: \{Entity\}\par{}Relation: \{List of Relations\}\par{}Answer:                                                                                                                                                                                                                                                                                                                                                                                                                                                                                                                                                                                                                                                                                                                                                                                                                                                                                                                                                                                                                                                                                                                                                                              \\ 
\hline
\textbf{Generate}\par{}\textbf{Triples}                                      & Given the existing triples, please generate some new triples related to your current thought. These new triples may come from your inherent knowledge directly or reasoning from the given triples.\par{}Here are some examples.\par{}\textit{In-Context Few-shot}\par{}Thought:~\{Thought\}\par{}Known Triples: \{Explored Triples\}\par{}Generated Triples:                                                                                                                                                                                                                                                                                                                                                                                                                                                                                                                                                                                                                                                                                                                                                                                                                                                                                                                                                                                                                                                                                                                                                   \\ 
\hline
\textbf{Verify}\par{}\textbf{Triples}                                        & Given the existing triples please select relevant triples to the question from LLM-generated triples based on your inherent knowledge.\par{}Here are some examples.\par{}\textit{In-Context Few-shot}Question: \{Question\}\par{}Generated triples: \{Generated triples\}\par{}Answers:                                                                                                                                                                                                                                                                                                                                                                                                                                                                                                                                                                                                                                                                                                                                                                                                                                                                                                                                                                                                                                                                                                                                                                                                                         \\
\hline
\end{tabular}
\caption{Prompts for different tasks used in GoG.}
\label{tab:prompts}
\end{table*}

\section{Case Study}
\label{app:case_study}

\subsection{Comparison between ToG and GoG under CKG setting}
\label{app:cmp_ckg}
ToG is likely to think compound value types (CVT) are not worthy to further explore and ignore them, as they do not offer information directly. Our GoG can easily solve this problem by expanding subgraph dynamically, that means if there is not enough information provided by the current subgraph, GoG would search one more hop, so the neighbors of CVT nodes is taken into consideration in this way. As illustrated in Table \ref{tab:cmp_ckg}, In this case, ToG gets lost and doesn't retrieve correct information when encounters CVT, "\textit{UnName\_Entity}" represents CVT nodes in the explored paths. On the contrast, our GoG can handle CVT nodes well by further searching.

\begin{table*}
\centering
\begin{tabular}{>{\hspace{0pt}}m{0.1\linewidth}|>{\hspace{0pt}}m{0.867\linewidth}} 
\hline
Question                & Where did the "Country Nation World Tour" concert artist go to college?                                                                                                                                                                                                                                                                                                                                                                                                                                                                                                                                                                                                                                                                                                                                                                                                                                                                                                                                                                                                                                                                                                                                                                                                                                                                                            \\ 
\hline
Dropped Crucial Triples & None                                                                                                                                                                                                                                                                                                                                                                                                                                                                                                                                                                                                                                                                                                                                                                                                                                                                                                                                                                                                                                                                                                                                                                                                                                                                                                                                                               \\ 
\hline
ToG                     & Reasoning Paths:~\par{}{[}(Country Nation World Tour,~music.artist.concert\_tours,~Brad Paisley), (Country Nation World Tour,~music.concert\_tour.artist,~Brad Paisley), (Bachelor's degree, education.educational\_degree.people\_with\_this\_degree, UnName\_Entity)],\par{}{[}(UnName\_Entity, education.education.institution, University of North Carolina at Chapel Hill), (UnName\_Entity, education.education.major\_field\_of\_study, Journalism), (UnName\_Entity, education.education.degree, Bachelor's degree)]\par\null\par{}Prediction:\par{}Based on the given knowledge triplets, the artist who had the "Country Nation World Tour" concert is Brad Paisley. Additionally, it is mentioned that Brad Paisley has a Bachelor's degree. The knowledge triplets also provide information about the institution he attended, which is the University of North Carolina at Chapel Hill. Therefore, the answer to the question is \{\textcolor{red}{University of North Carolina at Chapel Hill}\}.                                                                                                                                                                                                                                                                                                                                                    \\ 
\hline
GoG                     & Thought 1:~I need to find the artist who performed the "Country Nation World Tour" and then find out where they went to college.\par{}Action 1:~Search[Country Nation World Tour]\par{}Observation 1:~Country Nation World Tour, concert\_tour.artist, Brad Paisley\par\null\par{}Thought 2:~Brad Paisley performed the "Country Nation World Tour", so I need to find out where Brad Paisley went to college.\par{}Action 2:~Search[Brad Paisley]\par{}Observation 2: Brad Paisley, person.education, m.0h3d7qb\par{}Brad Paisley, person.education, m.0h3d7qj\par{}Brad Paisley, person.education, m.0n1dd\_6\par\null\par{}Thought 3:~m.0h3d7qb,~m.0h3d7qj and~m.0n1dd\_6 could be CVT, I need to further search them.\par{}Action 3: Search[m.0h3d7qb \textbar{} m.0h3d7qj \textbar{} m.0n1dd\_6]\par{}Observation 3:~Brad Paisley, person.education, m.0h3d7qj\par{}m.0h3d7qb, education.institution, John Marshall High School\par{}m.0h3d7qj, education.institution, Belmont University\par{}m.0h3d7qj, education.major\_field\_of\_study, Music\par{}m.0n1dd\_6, education.institution, West Liberty University\par\null\par{}Thought 4:~Based on the given observations, Brad Paisley went to Belmont University. This is where the "Country Nation World Tour" concert artist went to college.\par{}Action 4:~Finish[\textcolor{blue}{Belmont University}]  \\
\hline
\end{tabular}
\caption{The comparison between ToG and GoG under CKG setting. The \textcolor{red}{wrong answers} are highlighted with red color, and the \textcolor{blue}{correct answers} are highlighted with blue color.}
\label{tab:cmp_ckg}
\end{table*}

\subsection{Comparison between ToG and GoG under IKG setting}
\label{app:cmp_ikg}
In this section, we present a case analysis to evaluate the utility of GoG, as demonstrated in Table \ref{tab:cmp_ikg}. In this case, GoG will first search the neighbors of "Appalachian Mountains" and obtain \textbf{(Bald Eagle Mountain, mountain.mountain\_range, Appalachian Mountains)}, \textbf{(Spaulding Mountain, mountain.mountain\_range, Appalachian Mountains)} and
\textbf{(Old Rag Mountain, mountain.mountain\_range, Appalachian Mountains)}, then GoG can infer that Appalachian Mountains are also located in North America simply by knowing information about the other three mountains. However, ToG fails on this question once the crucial triple \textbf{(Appalachian Mountains, location.location.containedby, North America)} is missing. In another word, for one-hop questions, GoG can use the neighbors information of the topic entities to predict the tail entities while other methods can not make full use of the neighbors information.

\begin{table*}
\centering
\begin{tabular}{>{\hspace{0pt}}m{0.1\linewidth}|>{\hspace{0pt}}m{0.867\linewidth}} 
\hline
Question                & Where do the appalachian mountains run?                                                                                                                                                                                                                                                                                                                                                                                                                                                                                                                                                                                                                                                                                                                                                                                                                                                                                                                                                                                                                                                                                                                                                                                                                                \\ 
\hline
Dropped Crucial Triple~ & Appalachian Mountains, location.location.containedby, North America                                                                                                                                                                                                                                                                                                                                                                                                                                                                                                                                                                                                                                                                                                                                                                                                                                                                                                                                                                                                                                                                                                                                                                                                    \\ 
\hline
ToG                     & Reasoning Paths: None\par\null\par{}Prediction:~I'm sorry, but I need more information to answer your question.                                                                                                                                                                                                                                                                                                                                                                                                                                                                                                                                                                                                                                                                                                                                                                                                                                                                                                                                                                                                                                                                                                                                                        \\ 
\hline
GoG                     & Thought 1:~I need to search the Appalachian Mountains to find out where they run.\par{}Action 1:~Search[Appalachian Mountains]\par{}Observation 1:~Appalachian Mountains, location.contains, Brasstown Bald\par{}Appalachian Mountains, location.contains, Craggy Dome\par{}Appalachian Mountains, location.geolocation, m.02\_qbv6\par{}Bald Eagle Mountain, mountain.mountain\_range, Appalachian Mountains\par{}Spaulding Mountain, mountain.mountain\_range, Appalachian Mountains\par{}Old Rag Mountain, mountain.mountain\_range, Appalachian Mountains\par\null\par{}Thought 2: The Appalachian Mountains contain various locations and are part of geolocations. To find out where they run, I need to generate some new triples based on these given triples and my inherent knowledge.\par{}Action 2:~ Generate[where do the Appalachian Mountains run]\par{}Observation 2:~Appalachian Mountains, geographical\_feature.contains\_location, Blue Ridge Mountains\par{}Appalachian Mountains, geographical\_feature.start\_location, Alabama\par{}Appalachian Mountains, geographical\_feature.start\_location,~North America\par\null\par{}Thought 3:~The Appalachian Mountains start from~North America\par{}Action 3: Finish[Alabama \textbar{} \textcolor{blue}{North America}]  \\ 
\hline
Ground Truth            & North America                                                                                                                                                                                                                                                                                                                                                                                                                                                                                                                                                                                                                                                                                                                                                                                                                                                                                                                                                                                                                                                                                                                                                                                                                                                          \\
\hline
\end{tabular}
\caption{The comparison between ToG and GoG under IKG setting. The \textcolor{red}{wrong answers} are highlighted with red color, and the \textcolor{blue}{correct answers} are highlighted with blue color.}
\label{tab:cmp_ikg}
\end{table*}

\subsection{Cases of Generate Action}

Generate Action is typically invoked when GoG cannot directly obtain an answer from the search results. There are two possible reasons for not being able to directly get an answer from the search results: (1) The correct relation was not selected, or (2) There is a lack of knowledge (the corresponding triples have been deleted). 
For example, in the case of Table \ref{tab:gen-1}, GoG doesn't select the correct relation \textit{"base.biblioness.bibs\_location.state"} in \textit{Search[Montreal]}, which 
leads to the answer not appearing in the search results. At this point, GoG uses the internal knowledge of LLMs to generate an answer to the question \textit{ "which Canadian province is Montreal in" } and successfully gets the correct answer \textit{"Quebec"}. 

In the case of Table \ref{tab:gen-2}, the key triple \textit{"Guatemala, location.location.containedby, Central America"} is missing, which also leads to the information not being found in the search. At this point, GoG uses the internal knowledge of LLMs along with the retrieved triple \textit{"Guatemala, countries.continent, North America"} to successfully infer that Guatemala is located in Central America, thereby obtaining the correct answer.

\begin{table*}
\centering
\begin{tabular}{>{\hspace{0pt}}m{0.177\linewidth}|>{\hspace{0pt}}m{0.765\linewidth}} 
\hline
Question                & Which Canadian Province was the locatiuon for the TV show the Lottery?                                                                                                                                                                                                                                                                                                                                                                                                                                                                                                                                                                                                                                                                                                                                                                                                                                                                                                                                                                                                                                                                                                                                                                                                                                                                                                                                                                                                                                                                                                                                                                                                    \\ 
\hline
Dropped Crucial Triples & None                                                                                                                                                                                                                                                                                                                                                                                                                                                                                                                                                                                                                                                                                                                                                                                                                                                                                                                                                                                                                                                                                                                                                                                                                                                                                                                                                                                                                                                                                                                                                                                                                                                                      \\ 
\hline
GoG                     & Thought 1: I need to find the location of the TV show 'The Lottery'.\par{}Action 1: Search[The Lottery]\par{}Observation 1: The Lottery, tv\_program.country\_of\_origin, United States of America\par{}The Lottery, tv\_program.filming\_locations, Montreal\par{}The Lottery, tv\_program.original\_network, m.010pllm8\par\null\par{}Thought 2: The Lottery was filmed in Montreal, but I need to identify the Canadian province that Montreal is in.\par{}Action 2: Search[Montreal]\par{}Observation 2: Hochelaga-Maisonneuve, location.containedby, Montreal\par{}Les Bobards, location.containedby, Montreal\par{}Mercier, Montreal, location.containedby, Montreal\par{}Montreal, travel\_destination.tourist\_attractions, Canadian Centre for Architecture\par{}Montreal, travel\_destination.tourist\_attractions, Fantasia Film Festival\par{}Montreal, travel\_destination.tourist\_attractions, Marathon Oasis de Montreal\par{}Montreal, travel\_destination.tourist\_attractions, Montreal International Reggae Festival\par{}Montreal, travel\_destination.tourist\_attractions, Montreal Museum of Fine Arts\par{}New York Life Insurance Building, location.containedby, Montreal\par{}The Pirhana Bar, location.containedby, Montreal\par\null\par{}Thought 3: Montreal is a city, so I need to find the province it belongs to.\par{}Action 3: \textcolor[rgb]{0,0.502,0}{Generate[which Canadian province is Montreal in]}\par{}Observation 3: Montreal, location.located\_in, Quebec"\par{}Quebec, country\_part, Canada\par\null\par{}Thought 4: The answer is the province of Quebec in Canada.\par{}Action 4: Finish[\textcolor{blue}{Quebec}]  \\ 
\hline
Ground Truth            & Quebec                                                                                                                                                                                                                                                                                                                                                                                                                                                                                                                                                                                                                                                                                                                                                                                                                                                                                                                                                                                                                                                                                                                                                                                                                                                                                                                                                                                                                                                                                                                                                                                                                                                                    \\
\hline
\end{tabular}
\caption{Case for Generate Action in GoG under CKG setting. The \textcolor{blue}{correct answers} are highlighted with blue color. The \textcolor[rgb]{0,0.502,0}{Generate Action} is highlighted with green color.}
\label{tab:gen-1}
\end{table*}

\begin{table*}
\centering
\begin{tabular}{>{\hspace{0pt}}m{0.094\linewidth}|>{\hspace{0pt}}m{0.848\linewidth}} 
\hline
Question                & Which nation has the Alta Verapaz Department and is in Central America?                                                                                                                                                                                                                                                                                                                                                                                                                                                                                                                                                                                                                                                                                                                                                                                                                                                                                                                                                                                                                                                                                                                                                                                                                                                                                                                                                                                                                                                                                                                                                                                                                                                                                                                                                                                                                                                                                                                            \\ 
\hline
Dropped Crucial Triples & Alta Verapaz Department, location.location.containedby, Guatemala\par{}Guatemala, location.location.contains, Alta Verapaz Department\par{}Guatemala, location.country.administrative\_divisions, Alta Verapaz Department\par{}Alta Verapaz Department, location.administrative\_division.country, Guatemala\par{}Central America, location.location.contains, Guatemala\par{}Guatemala, location.location.containedby, Central America                                                                                                                                                                                                                                                                                                                                                                                                                                                                                                                                                                                                                                                                                                                                                                                                                                                                                                                                                                                                                                                                                                                                                                                                                                                                                                                                                                                                                                                                                                                                                            \\ 
\hline
GoG                     & Thought 1: I need to find the nation that contains the Alta Verapaz Department.\par{}Action 1: Search[Alta Verapaz Department]\par{}Observation 1: Alta Verapaz Department, administrative\_area.administrative\_area\_type, Guatemalan department\par{}Alta Verapaz Department, statistical\_region.population, g.11bv5vs59b\par{}Alta Verapaz Department, statistical\_region.population, m.04q66h7\par{}Raxruha, location.containedby, Alta Verapaz Department\par{}San Agust\u00edn Lanqu\u00edn, location.containedby, Alta Verapaz Department\par{}San Pedro Carch\u00e1, location.containedby, Alta Verapaz Department\par{}Santa Catalina la Tinta, location.containedby, Alta Verapaz Department\par{}Santa Mar\u00eda Cahab\u00f3n, location.containedby, Alta Verapaz Department\par\null\par{}Thought 2: I found that Alta Verapaz Department is a Guatemalan department, now I need to confirm if Guatemala is in Central America.\par{}Action 2: Search[Guatemala]\par{}Observation 2: Guatemala, countries.continent, North America\par{}Guatemala, country.languages\_spoken, Spanish Language\par\null\par{}Thought 3: The information says Guatemala is in North America, not Central America. This may be due to the definition of \"Central America\" not being strictly a continent but a region. I need to verify if Guatemala is considered a part of Central America based on the given topic entity.\par{}Action 3: \textcolor[rgb]{0,0.502,0}{Generate[Central America includes Guatemala]}\par{}Observation 3: Alta Verapaz Department, country.administrative\_area\_of, Guatemala\par{}Central America, region.includes, Guatemala\par{}Guatemala, country.geographic\_region, Central America\par{}North America, continent.includes, Central America\par\null\par{}Thought 4: Based on the new triples, I can confirm that Guatemala, which contains the Alta Verapaz Department, is indeed in Central America.\par{}Action 4: Finish[\textcolor{blue}{Guatemala}]  \\ 
\hline
Ground Truth            & Guatemala                                                                                                                                                                                                                                                                                                                                                                                                                                                                                                                                                                                                                                                                                                                                                                                                                                                                                                                                                                                                                                                                                                                                                                                                                                                                                                                                                                                                                                                                                                                                                                                                                                                                                                                                                                                                                                                                                                                                                                                          \\
\hline
\end{tabular}
\caption{Case for Generate Action in GoG under IKG setting. The \textcolor{blue}{correct answers} are highlighted with blue color. The \textcolor[rgb]{0,0.502,0}{Generate Action} is highlighted with green color.}
\label{tab:gen-2}
\end{table*}

\subsection{Bad Cases of GoG}

\textbf{Hallucination}

The case in Table \ref{tab:hallucination} demonstrates an  example of GoG, where GoG mistakenly inferred that the date "2012-01-01" was when the team won the championship, while in fact, this date marks the beginning of Larr Baer's leadership of the team. 

\begin{table*}
\centering
\begin{tabular}{>{\hspace{0pt}}m{0.171\linewidth}|>{\hspace{0pt}}m{0.771\linewidth}} 
\hline
Question                & When was the last time a team led by Larr Baer win the World Series?                                                                                                                                                                                                                                                                                                                                                                                                                                                                                                                                                                                                                                                                                              \\ 
\hline
Dropped Crucial Triples & None                                                                                                                                                                                                                                                                                                                                                                                                                                                                                                                                                                                                                                                                                                                                                              \\ 
\hline
GoG                     & Thought 1:~I need to find the team led by Larry Baer and then find out when they last won the World Series. college.\par{}Action 1:~Search[Larry Baer]\par{}Observation 1:~Larry Baer, board\_member.leader\_of, m.011zsc4\_\par\null\par{}Thought 2:~I need to find out the team led by Larry Baer and then find out when they last won the World Series.\par{}Action 2:~Search[m.011zsc4\_]\par{}Observation 2: Larry Baer, board\_member.leader\_of, m.011zsc4\_\par{}San Francisco Giants, organization.leadership, m.011zsc4\_\par{}m.011zsc4\_, leadership.from, 2012-01-01\par\null\par{}Thought 3:~The San Francisco Giants is the team led by Larry Baer and they last won the World Series in 2012.\par{}Action 3: Finish[\textcolor{red}{2012-01-01}]  \\ 
\hline
Ground Truth            & 2014 World Series                                                                                                                                                                                                                                                                                                                                                                                                                                                                                                                                                                                                                                                                                                                                                 \\
\hline
\end{tabular}
\caption{Case for hallucination in GoG. The \textcolor{red}{wrong answers} are highlighted with red color.}
\label{tab:hallucination}
\end{table*}

\noindent \textbf{Generation Error}

The case in Table \ref{tab:generate_error} demonstrates an  example of GoG, where GoG make an error in generating the mascot for Syracuse University Athletics. It mistakenly identified "Orangeman" as the mascot of the team, but in reality, the team's mascot is Otto the Orange. "Orangeman" is actually the name used to refer to the team.

\begin{table*}
\centering
\begin{tabular}{>{\hspace{0pt}}m{0.127\linewidth}|>{\hspace{0pt}}m{0.815\linewidth}} 
\hline
Question                & What is Syracuse University Athletic's mascot?                                                                                                                                                                                                                                                                                                                                                                                                                                                                                                                                                                                                                                                                                                                                                                                                                                                                                                          \\ 
\hline
Dropped Crucial Triples & None                                                                                                                                                                                                                                                                                                                                                                                                                                                                                                                                                                                                                                                                                                                                                                                                                                                                                                                                                    \\ 
\hline
GoG                     & Thought 1:~I need to find the mascot of Syracuse University Athletics."\par{}Action 1:~Search[Syracuse University Athletics]\par{}Observation 1:~Syracuse University Athletics, organization.headquarters, m.0\_gdsbw\par{}Syracuse University Athletics, organization.parent, m.0105kgwd\par{}m.0\_gdrcn, award\_honor.award\_winner, Syracuse University Athletics\par{}m.0\_y3csk, award\_honor.award\_winner, Syracuse University Athletics\par\null\par{}Thought 2:~There is no information about the mascot of Syracuse University Athletics, so I need to generate more triples based on these given triples and my inherent knowledge.\par{}Action 2:~Generate[What is the mascot of Syracuse University Athletics]\par{}Observation 2: Syracuse University Athletics, athletics\_brand.mascot, Orangemen\par\null\par{}Thought 3:~The mascot of Syracuse University Athletics is Orangemen.\par{}Action 3: Finish[\textcolor{red}{Orangemen}]  \\ 
\hline
Ground Truth            & Otto the Orange                                                                                                                                                                                                                                                                                                                                                                                                                                                                                                                                                                                                                                                                                                                                                                                                                                                                                                                                         \\
\hline
\end{tabular}
\caption{Case for generation error in GoG. The \textcolor{red}{wrong answers} are highlighted with red color.}
\label{tab:generate_error}
\end{table*}

\noindent \textbf{Decompose Error}

The case in Table \ref{tab:decompose_error} demonstrates an  example of GoG. In this example, during the process of decomposing the problem, GoG forgets that the initial goal is to find the team coached by Pablo Laso. Instead, in the final thought, the objective shifts to finding the country where this team is located. This type of situation is likely to become more frequent as the number of search iterations increases and the context length grows longer.

\begin{table*}
\centering
\begin{tabular}{>{\hspace{0pt}}m{0.148\linewidth}|>{\hspace{0pt}}m{0.794\linewidth}} 
\hline
Question                & What basketball team in Spain is Pablo Laso head coach for?                                                                                                                                                                                                                                                                                                                                                                                                                                                                                                                                                                                                                                                                                                                                                                                                                                                                                                                                                                                                                                                                         \\ 
\hline
Dropped Crucial Triples & None                                                                                                                                                                                                                                                                                                                                                                                                                                                                                                                                                                                                                                                                                                                                                                                                                                                                                                                                                                                                                                                                                                                                \\ 
\hline
GoG                     & Thought 1:~I need to find the basketball team that Pablo Laso is the head coach for in Spain.\par{}Action 1:~Search[Pablo Laso]\par{}Observation 1:~Pablo Laso, basketball\_coach.team, Real Madrid Baloncesto\par{}Pablo Laso, pro\_athlete.teams, m.012n525v\par{}Real Madrid Baloncesto, basketball\_team.head\_coach, Pablo Laso\par\null\par{}Thought 2:~ Pablo Laso is the head coach for Real Madrid Baloncesto, so I need to find out which country Real Madrid Baloncesto is located in.\par{}Action 2:~Search[Real Madrid Baloncesto]\par{}Observation 2:~Real Madrid Baloncesto, sports\_team.arena\_stadium, Palacio de Deportes de la Comunidad de Madrid\par{}Real Madrid Baloncesto, sports\_team.arena\_stadium, Raimundo Saporta Pavilion\par{}Real Madrid Baloncesto, sports\_team.location, Madrid\par{}Real Madrid Baloncesto, sports\_team.location, Spain\par{}Real Madrid Baloncesto, sports\_team.venue, m.0wz2\_rd\par{}Real Madrid Baloncesto, sports\_team.venue, m.0wz2wth\par\null\par{}Thought 3:~~Real Madrid Baloncesto is located in Madrid, Spain.\par{}Action 3:~Finish[\textcolor{red}{Spain}]  \\ 
\hline
Ground Truth            & Real Madrid Baloncesto                                                                                                                                                                                                                                                                                                                                                                                                                                                                                                                                                                                                                                                                                                                                                                                                                                                                                                                                                                                                                                                                                                              \\
\hline
\end{tabular}
\caption{Case for decompose error in GoG. The \textcolor{red}{wrong answers} are highlighted with red color.}
\label{tab:decompose_error}
\end{table*}

\end{document}